\documentclass{new_tlp}

\submitted{TBD}
\revised{TBD}
\accepted{TBD}

\usepackage{mathptmx}
                  
\usepackage{csquotes}
\usepackage{url}
\usepackage{amsmath}
\usepackage{pgf}

\usepackage[capitalize]{cleveref}

\newcommand{\slv}[1]{\textsc{#1}} 
\newcommand{\alphaslv}{\slv{Alpha}}
\newcommand{\clingo}{\slv{Clingo}}
\newcommand{\omiga}{\slv{OMiGA}}
\newcommand{\asperix}{\slv{ASPeRiX}}
\newcommand{\lazywasp}{\slv{Lazy Wasp}}

\newcommand{\prog}{\ensuremath{P}}

\newcommand{\assignment}{\ensuremath{A}}

\newcommand{\assignmentp}{\ensuremath{A}^+}
\newcommand{\assignmentm}{\ensuremath{A}^-}

\newcommand{\preds}{\ensuremath{\mathcal{P}}}
\newcommand{\vars}{\ensuremath{\mathrm{vars}}}
\newcommand{\nafsymbol}{\ensuremath{\mathrm{not}}}
\newcommand{\naf}[1]{\ensuremath{\nafsymbol~#1}}
\newcommand{\head}{\ensuremath{\mathrm{H}}}
\newcommand{\body}{\ensuremath{\mathrm{B}}}
\newcommand{\bodyp}{\ensuremath{\mathrm{B}^+}}
\newcommand{\bodyn}{\ensuremath{\mathrm{B}^-}}
\newcommand{\ground}{\ensuremath{\mathrm{grd}}}

\newcommand{\gratoms}{\ensuremath{\mathit{At}_{\mathrm{grd}}}}
\newcommand{\consts}{\ensuremath{\mathcal{C}}}
\newcommand{\variables}{\ensuremath{\mathcal{V}}}

  \title[Advancing Lazy-Grounding ASP Solving Techniques]
        {Advancing Lazy-Grounding ASP Solving Techniques -- Restarts, Phase Saving, Heuristics, and More}

  \author[A.\ Weinzierl, R.\ Taupe, G.\ Friedrich]
  	{
         ANTONIUS WEINZIERL\textsuperscript{1}, RICHARD TAUPE\textsuperscript{2,3}, and GERHARD FRIEDRICH\textsuperscript{2}\\
         \textsuperscript{1} TU Wien (Vienna University of Technology), Austria,\\
         \email{antonius.weinzierl@kr.tuwien.ac.at}\\
         \textsuperscript{2} Alpen-Adria-Universität, Klagenfurt, Austria,\\
         \email{gerhard.friedrich@aau.at}\\
         \textsuperscript{3} Siemens AG Österreich,\\
         \email{richard.taupe@siemens.com}
     }

\jdate{TBD}
\pubyear{TBD}
\pagerange{\pageref{firstpage}--\pageref{lastpage}}
\doi{TBD}

\begin{document}

\label{firstpage}

\maketitle

\begin{abstract}
	Answer-Set Programming (ASP) is a powerful and expressive knowledge
	representation paradigm with a significant number of applications in logic-based AI. The
	traditional ground-and-solve approach, however, requires ASP programs to be
	grounded upfront and thus suffers from the so-called grounding bottleneck
	(i.e., ASP programs easily exhaust all available memory and thus become
	unsolvable). As a remedy, lazy-grounding ASP solvers have been developed,
	but many state-of-the-art techniques for grounded ASP solving have not been
	available to them yet. In this work we present, for the first time, adaptions
	to the lazy-grounding setting for many important techniques, like restarts,
	phase saving, domain-independent heuristics, and learned-clause deletion.
	Furthermore, we investigate their effects and in general observe a large
	improvement in solving capabilities and also uncover negative effects in
	certain cases, indicating the need for portfolio solving as known from other
	solvers. \textit{Under consideration for acceptance in TPLP.}
\end{abstract}

  \begin{keywords}
    Answer-Set Programming,
    Lazy-Grounding,
    Solving Techniques,
    Knowledge Representation
  \end{keywords}

\section{Introduction}
Answer-Set Programming is employed in many application areas
\cite{DBLP:journals/ki/FalknerFSTT18} because ASP offers a rich first-order
declarative knowledge representation language, and powerful reasoning systems
are available. For hard, practical configuration problems such as the Partner
Units Problem
\cite{DBLP:conf/cpaior/AschingerDFGJRT11,DBLP:journals/ai/Teppan17}, for
example, ASP was applied successfully off-the-shelf. However, there are
practical problem instances in configuration, scheduling, and planning, where
pure ASP systems based on the traditional ground-and-solve approach cannot
compute solutions because of excessive main-memory consumption in the
grounding phase, which
is frequently superlinear in the size of the input
\cite{DBLP:journals/ki/FalknerFSTT18}.

One way to tackle the grounding issue is by grounding lazily only those parts
of a first-order theory which are actually needed to solve the problem at
hand. This lazy grounding is a bottom-up procedure that interleaves grounding
and solving in such a way that parts of the grounding are constructed when the
solver needs them. There exist a number of lazy-grounding ASP solvers, GASP
\cite{gasp}, Omiga \cite{omiga_system}, ASPeRiX \cite{Lefevre.2017} and the
recently introduced \alphaslv\ \cite{alpha_technical}. Only the latter
integrates lazy grounding with a conflict-driven clause-learning (CDCL)
solver, hence it currently is the most efficient lazy-grounding system for ASP
solving.

Nevertheless, \alphaslv\ only realizes a subset
(cf.~\citeNP{alpha_technical}; \citeNP{DBLP:conf/inap/LeutgebW17}) of the techniques
usually employed in ground-and-solve ASP systems, whose efficiency is largely
due to their use of a wide range of CDCL techniques for efficient SAT solving
\cite{clasp_journal,DBLP:conf/lpnmr/AlvianoCDFLPRVZ17}. Thus the search
performance of \alphaslv\ is significantly worse than that of ground-and-solve systems
on problems where grounding itself is not an issue. Lazy grounding at its core
contains some specific restrictions (e.g.~guessing on all atoms is not
allowed) that are of no concern for the techniques employed in
ground-and-solve systems. Hence, one cannot simply add lazy grounding on top of
the existing solving techniques. Quite the contrary, each technique from
ground-and-solve systems must be checked for suitability to the lazy-grounding
setting individually.
In particular, restarts, phase saving, domain-independent heuristics, and
learned-clause deletion, which are crucial methods in grounded ASP solving to
deal with hard problem instances, are not available to lazy-grounding
solvers.

In this work we show how these methods must be enhanced to enable efficient
search for answer sets based on lazy-grounding ASP solving. Our contributions
are as follows.
\begin{itemize}
\item An investigation into the techniques of restarts, phase saving,
  domain-independent heuristics, and learned-clause deletion, determining
  compatibility with lazy grounding.
\item Enhancing these methods to fit the lazy-grounding setting, investigating
  their effects and creating novel adaptions to work around issues specific to
  lazy grounding.
\item Specifically, the introduction of domain-independent VSIDS-like
  heuristics that use atom-dependency information to assign atom-activity
  scores to ground rules.
\item An integration of the enhanced methods in the latest version of \alphaslv.
\item An evaluation of the above methods on worst-case scenarios for lazy
  grounding (and \alphaslv), where grounding is easy but problem solving is
  challenging. Our evaluations show signifiant runtime improvements, i.e.\ up
  to a factor of three. Furthermore, our experiments also indicate that the novel
  techniques introduce no obstacle for solving instances that are hard to ground.
\end{itemize}

This paper starts with an introduction of the basics of ASP in
Section~\ref{sec:preliminaries}. In Section~\ref{sec:techniques} we recap the
principles of several state-of-the-art techniques for grounded ASP solving and
show their enhancements and the adaptions required for lazy-grounding systems. The
runtime improvements of these new techniques are exemplified in
Section~\ref{sec:experiments}. Section~\ref{sec:related} discusses related
work and Section~\ref{sec:conclusion} concludes.

\section{Preliminaries}
\label{sec:preliminaries}

Let $\consts$ be a finite set of constants, $\variables$ be a set of variables and $\preds$ be a finite set of predicates.
An atom is an expression $p(t_1,\dots,t_n)$ where $p$ is an $n$-ary predicate and $t_1,\dots,t_n \in \consts \cup \variables$ are terms, and a literal is either an atom $a$ or its default negation $\naf{a}$.
An ASP program \prog\ is a finite set of (normal) rules of the form\\[0.5ex]
\centerline{$h \leftarrow b_1,~\ldots,~b_m,~\naf{b_{m+1}},~\ldots,~\naf{b_n}.$}\\[0.5ex]
where $h$ and $b_1,\dots,b_m$ are positive literals (i.e.\ atoms) and $\naf{b_{m+1}}$, $\ldots$, $\naf{b_n}$ are negative literals.
Given a rule $r$, we denote by $\head(r)=\{h\}$, $\body(r) = \{ b_1,\dots,b_m,$ $\naf{b_{m+1}},$
$\ldots,\naf{b_n} \}$, 
$\bodyp(r) = \{ b_1, \dots, b_m \}$, and 
$\bodyn(r) = \{ b_{m+1},$ $\dots,$ $b_n \}$ 
the head, the body, the positive body, and the negative body of $r$, respectively.
If $\head(r) = \emptyset$, $r$ is a called a constraint, and a fact if $\body(r) = \emptyset$.
Given a literal $l$, set of literals $L$, or rule $r$, we denote by $\vars(l)$, $\vars(L)$, or $\vars(r)$ the set of variables occurring in $l$, $L$, or $r$, respectively. 
A literal $l$ or rule $r$ is ground if $\vars(l) = \emptyset$ or $\vars(r) =
\emptyset$, respectively.  The set of all ground atoms is denoted by
$\gratoms$. A program $\prog$ is ground if all its rules $r \in \prog$ are ground. 

An interpretation $I \subseteq \gratoms$
satisfies a ground rule $r$, denoted $I \models r$, if
$\bodyp(r) \subseteq I \land \bodyn(r) \cap I = \emptyset$ implies
$\head(r) \subseteq I$ and $\head(r) \neq \emptyset$.
$I$ is an \emph{answer set} of a ground program $\prog$ if $I$ is the subset-minimal model of $\prog^I$, where
$\prog^I = \{ r \in \prog \mid \bodyp(r) \subseteq I \land \bodyn(r) \cap I = \emptyset \}$
is the so-called FLP reduct, the set of rules whose body is satisfied by $I$ \cite{DBLP:journals/ai/FaberPL11}.
A (partial) \emph{assignment} $\assignment$ is a set of signed atoms where $\assignmentp$ denotes the atoms assigned a positive truth value and $\assignmentm$ those assigned a negative truth value in $\assignment$.
Given an atom $\mathit{at}$ the result of applying a substitution
$\sigma: \variables \to \consts$ to $\mathit{at}$ is denoted by $\mathit{at}\sigma$; this is extended in the
usual way to rules $r$, i.e., $r\sigma$ for a rule of the above form is
$h\sigma \leftarrow b_1\sigma, \ldots, b_m\sigma, \naf b_{m+1}\sigma, \naf
b_n\sigma$. The grounding of a rule is given by
$\ground(r) = \{ r\sigma \mid \sigma \text{ is a substitution for all }v \in
\vars(r)\}$ and the grounding $\ground(\prog)$ of a program $\prog$ is given by
$\ground(\prog) = \bigcup_{r \in \prog} \ground(r)$.
The answer sets of a non-ground program $\prog$ are given by the answer sets of $\ground(\prog)$.

Lazy grounding is an approach to tackle the \textit{grounding bottleneck} inherent in traditional ground-and-solve systems which makes programs whose grounding exceeds available memory unsolvable.
We describe lazy grounding only briefly here and 
refer to \citeN{alpha_technical} and \citeN{DBLP:conf/inap/LeutgebW17} for a detailed account of the lazy-grounding ASP system \alphaslv.
Computing all answer sets such that $\ground(\prog)$ is constructed lazily is typically done by a loop composed of two phases: given a partial assignment (that is initially empty), first ground those rules that potentially fire under the current assignment, second expand the current assignment (using propagation and guessing). If the loop reaches a fixpoint, i.e., no more rules potentially fire and nothing is left to propagate or guess on, and no constraints are violated, then the current assignment is an answer set.

One important difference to ground-and-solve is that a lazy-grounding solver does not guess on each atom whether it is true or
false, but it guesses about ground instances of rules whether they fire or not.
This is correct due to the underlying solving mechanisms based on computation sequences and avoids generating completion nogoods (cf.\ \citeNP{DBLP:conf/adbt/Clark77}; \citeNP{clasp_journal}) in most cases.
This has the advantage that less space is occupied by completion nogoods and the drawback that the solver lacks information, like in the following example:\\[0.5ex]
\centerline{$p(X) \leftarrow q(X,Y).$}\\[0.5ex]
If the solver knows that, say, $p(13)$ must be true, in lazy grounding the solver generally does not know if there are any rules that can derive $p(13)$ but have not yet been grounded, and therefore cannot conclude that one of these rules must fire.

\section{State-of-the-Art Solving Techniques}
\label{sec:techniques}

In the following we discuss several important state-of-the-art techniques for
ASP solving and investigate their adaption to the lazy-grounding
setting. Since many of the techniques for efficient ASP solving originate in
SAT solving, there are many similarities to SAT techniques. As mentioned in
the previous section, lazy-grounding ASP solving imposes additional
restrictions that are neither considered in SAT solving nor in the traditional
approach for ASP solving, like the fact that not all ground atoms are known
from the beginning. Hence one cannot just put lazy grounding on top of the
existing technologies, but each technology must be individually checked for compatibility
with the lazy-grounding approach.
\citeN{DBLP:conf/aaai/BomansonJW19}, for example, uncovered that
supporting aggregates with a lazy-grounding ASP solver requires a sequential
enumeration of all ground terms that will appear during the run of the
solver. In the ground-and-solve approach this enumeration is trivially found
by simply looking at the full grounding of the input program, while in
lazy grounding the solver must provide special facilities to enable an
efficient enumeration that does not rely on knowing all ground rules in
advance. Similarly, while virtually all ground-and-solve systems do a Clark's
completion to represent rules by clauses (or nogoods), in the lazy-grounding
approach the full completion cannot be obtained in advance without grounding
the input program fully. \citeN{DBLP:conf/ijcai/BogaertsW18} have developed on-demand
computation of justifications in order to get around the
issues of missing knowledge from the Clark's completion. Luckily, the
techniques we investigate here turned out to be rather well-behaved, requiring
less adaptions. Nevertheless, there were still some surprising challenges
we had to overcome.

\subsection{Restarts}
\label{sec:techniques:restarts}

Restarts originate in the observation of SAT solvers exhibiting heavy-tailed
behaviour, i.e., when solving a set of SAT instances, the majority of the time
is consumed by a relatively small number of instances that often time out
while the majority of instances are solved in relatively little time. For those
instances with a long run time, the search seemingly gets stuck in some part of
the search space and restarting the search in a new part of the search space
helps \cite{DBLP:journals/jar/GomesSCK00}. Upon restarting all decisions
of the solver are undone, i.e., restarts are a backjump to decision level 0.
Importantly, restarts do not discard learned clauses and they do not reset
the search heuristics, i.e., highly active atoms (cf.\ \cref{sec:techniques:heuristics} for details)
before the restart are still considered by the heuristics as highly active
after the restart.

Even though restarts provide the solver with the possibility to go into a
completely different part of the search space after a restart, practical
experience showed that this is not optimal as it is likely that the search
gets stuck there again, while all recent knowledge (e.g.\ learned clasues) has
become mostly useless in the new search area.  Therefore, restarts are even
more efficient if combined with phase saving (cf.\ \cref{sec:techniques:phasesaving}), which leads the
solver again into the same area of the search space as it was before. This has
the important effect that learned clauses are still useful while the restart
effectively re-orders the sequence in which atoms have been guessed: after a
restart highly active atoms are chosen first, which leads to conflicts arising
much earlier than before the restart. Intuitively the binary search (sub-)tree
with conflicts at each of its leaves is now much more shallow as it contains fewer
irrelevant choices. These kind of restarts are very effective in uncovering
implicit information about the problem instance and restarts
in rapid succession significantly improve solving efficiency.

There are two principled ways for solvers to restart: static restart sequences
trigger a restart after a fixed number of conflicts, while adaptive (or dynamic) restarts
trigger a restart whenever the solver detects that it is not learning useful
clauses. \emph{Static restart sequences} often use the so-called Luby sequence
(cf.~\citeNP{DBLP:journals/ipl/LubySZ93}), while adaptive restarts
(cf.~\citeNP{DBLP:conf/cp/AudemardS12}) issue a restart not in a fixed sequence
depending on the number of conflicts encountered by the solver, but measure
the quality of learned clauses to decide whether a restart is
appropriate. \emph{Adaptive restarts} use the LBD (Literals Blocks Distance)
measure for quality of learned clauses
(cf.~\citeNP{DBLP:conf/ijcai/AudemardS09}), which intuitively counts the number
of decision levels of the literals appearing in the clause at the time the
clause is learned. The lower the LBD value, the better the clause will likely
perform in the remainder of the search. Adaptive restarts now compare an
average of the LBD value of learned clauses for the recent conflicts with the
average for the entire search to that point, and issue a restart if the recently
learned clauses have significantly worse LBD than the average of the whole
run. Computing these moving averages was later improved by
\citeN{DBLP:conf/sat/BiereF18} to consider \emph{exponential moving averages},
which allow evaluation without needing to queue all recently learned LBD
values.

\paragraph{Restarts for lazy grounding.}
\alphaslv\ now combines both restart strategies: adaptive restarts (which
usually are quick to trigger a restart) and static restart sequences that
allow exponentially increasing runs where no restart is triggered. The static
restart sequence is a Luby-sequence which is computed quickly by
\emph{reluctant doubling} (proposed by Donald E.~Knuth in his SAT'12 talk),
the state-of-the-art in most SAT solvers now. For adaptive restarts \alphaslv\
uses the exponential moving averages on LBD values as described above. Note
that \alphaslv\ currently does not update LBD values, as some SAT solvers do.

Since \alphaslv\ follows the original computation sequence
(cf.~\citeNP{Lefevre.2017}) for lazy-grounding answer-set computation,
picking atoms for guessing is restricted. First, only atoms that
represent the body of a rule with negation can be valid choice
points\footnote{This alone is known to be not optimal on certain input
  programs, cf.~\citeN{DBLP:conf/ecai/AngerGJS06}.},
and second, from these valid choice points only those
where the positive body of the rule is already derived may be picked for guessing
(cf.~\citeN{alpha_technical} for details). This severly \emph{restricts the
  order} in which atoms are chosen by the solver, i.e., it may forbid the
solver from branching on the most active atom(s). As a consequence of
that, restarts are less effective for the current lazy-grounding approach than
they are for ground-and-solve ASP systems. 

Luckily, for many search problems this negative effect does not manifest,
namely those where all potential choices are available to the solver right
from the beginning. An example of such a problem is graph colouring where each
choice point colors a vertex of the input graph. After a restart all choice
points are valid, hence a restart is indeed re-ordering atoms as in the
ground-and-solve cases.

For problems where guesses are ``stacked'', however, the
negative effects of restarting are visible. Examples of that are planning
problems where a choice on the second action may become valid only after the
first action was chosen. Clever reformulation of the problem may avoid
that issue, but this is beyond the scope of this work.

\subsection{Phase Saving}
\label{sec:techniques:phasesaving}

Phase saving (or progress saving) is a technique that focuses the search on a
specific part of the search space even after backjumping or restarts. Phase
saving means to save the \emph{last assigned value} of each atom and whenever
a choice is to be made on any atom, its last assigned value is taken
(cf.~\citeNP{DBLP:conf/sat/PipatsrisawatD07}). It does not matter if the last
value was assigned due to a choice or propagation. The effect of phase saving
is that when the solver is backjumping or restarting, the search is again
approaching the same point in the search space (i.e., the same candidate
answer set). The effect of phase saving alone seems to be less significant,
but in conjunction with restarts it has a tremenduous impact on performance
(cf.~\citeNP{DBLP:conf/ijcai/ElffersGGNS18}). It effectively makes the solver
approach the same point in the search space but from another direction and
leads to the new perspective of a solver as ``clause generating
machinery''. With the combination of phase saving and restarts the
uncovered clauses are small in size and pertinent to a small portion of the
search space, hence much more focused than without those techniques.

\paragraph{Phase saving for lazy grounding.}
Phase saving can be adapted to lazy grounding 
by adding an
array that keeps, for each known ground atom, its last assigned truth
value. Specific to lazy grounding, this array needs to \emph{grow} in size
during the run of the solver as a lazy-grounding solver uncovers ground atoms
step by step. 
We observed that the \emph{initial value} for
phases has a significant impact on whether an instance can be solved or not.
Note that if an atom is guessed whose truth value already is must-be-true,
then the phase is not considered but true is chosen directly, as otherwise a
conflict would arise immediately.

Our experiments include several settings for the initial phase: \emph{all
  false}, \emph{all true}, and \emph{random}. Selecting one or the other makes
a difference depending on the instance and it seems unlikely that one or the
other is always best. The experiments show however that all true seems to be slightly
more favourable though. The initially all false setting corresponds to the
MiniSat setting while the initially all true setting effectively corresponds
to what \clingo\ is doing. Note that \clingo\ actually uses true only for atoms
representing rule bodies, but since \alphaslv\ only guesses on atoms representing
rule bodies, it coincides with the all true setting.

\subsection{Domain-Independent Heuristics}
\label{sec:techniques:heuristics}

Heuristics for answer-set solving can roughly be classified as follows:
\textit{domain-independent heuristics} do not take the nature of the problem at hand into account, whereas \textit{domain-specific heuristics} have to be tailored to a specific problem.
Domain-specific heuristics are covered by
\citeN{DBLP:journals/corr/abs-1909-08231}, accordingly we focus on domain-independent heuristics in this work.
VSIDS (Variable State Independent Decaying Sum) \cite{Moskewicz.2001} and
BerkMin \cite{Goldberg.2002} are prominent domain-independent heuristics
originally developed for SAT but also successfully employed for ASP solving
(in \slv{clasp} \cite{clasp_journal} and \slv{wasp} \cite{Alviano.2013b}).

They assign a so-called \textit{activity} to every atom that counts the
number of times a clause containing this atom contributed to a conflict.
The activity of each atom is periodically divided by a constant (i.e., it
is \enquote{decayed}) to reduce the influence of conflicts further in the past.
When asked for an atom, the heuristics choose the most active one.
Other counters are maintained as well to enable the choice of which truth value to assign.
BerkMin additionally organizes the set of conflict clauses as a chronologically ordered stack, thereby preferring atoms in recent conflicts.
This is done to have regard to the fact that the set of atoms responsible for conflicts may change very quickly.

Atom activities are typically initialized by MOMs (Maximum Occurrences in clauses of Minimum size) \cite{DBLP:conf/aaai/GebserKROSW13,Pretolani.1993}.
A MOMs score for an atom is an estimate to what extent other atoms are affected when this atom is assigned.
For each atom, the MOMs score is a function of the number of nogoods involving the atom in a positive literal and the number of nogoods involving the atom in a negative literal.

A direct application of BerkMin or VSIDS to a lazy-grounding ASP solver like \alphaslv\ is challenging, because such a solver differs in many important ways from a solver adhering to the classical ground-and-solve paradigm.
One major difference is that not all ground rules, and consequently not all ground literals and atoms, are known at any given time to a lazy-grounding solver.
Because of this, a heuristic applied to lazy grounding can only incorporate atoms that are already known to the solver.

Another major difference lies in the solving mechanism:
while a traditional ASP solver can choose any atom to guess on, \alphaslv\ is
restricted to atoms representing rule bodies.
In other words, \alphaslv\ only guesses whether a certain rule fires or not, but it does not guess whether an atom in a rule's head or body is true or not.
A direct application of BerkMin or VSIDS to \alphaslv\ would therefore suffer from
the fact that choice points comprise only a small portion of all the literals
occuring in clauses (or nogoods) and therefore do not influence activity and
sign counters as much as other atoms.

\paragraph{Heuristics for lazy grounding.}
Domain-independent heuristics for lazy-grounding ASP solving were first
studied by \citeN{alpha_heuristics_paoasp}. The so-called class of
\textit{dependency-driven} heuristics has proven particularly useful and has
been further improved since. The basic idea is as follows: since for an
ordinary atom $b$ its truth cannot be guessed, find all choice points $c$ (i.e.,
atoms representing rule bodies) that have an influence on the truth of $b$ and
whenever the activity of $b$ is to be increased, increase the activity of $c$
instead. By that, choices are not done directly on highly active atoms but on
atoms that have an influence on highly active atoms, i.e., the solver is
focused on (ordinary) active atoms and chooses their truth value indirectly.
For an atom $b$ there are two ways how the guessing alone may influence its
value: either by firing a rule $r$ with $\head(r) = b$ or by firing a rule
$r'$ with $b \in \bodyn(r')$, since the firing of a rule makes all the atoms
in the negative body false and the head true.\footnote{Note
  that $\bodyp(r)$ is not affected because those atoms already have to
  be true for the rule to be a valid choice point. Also, $\head(r)$ is not necessarily
  affected if the rule is guessed to not fire, because there might be other rules
  with the same head, and for the atoms in $\bodyn(r)$ it is only known
  that one of them must be true to conform the rule not firing.}

\alphaslv's \emph{dependency-driven VSIDS} implementation maintains choice points in a heap data structure, which enables efficient access to the choice point with the highest activity.
After choosing an applicable choice point, the sign is chosen by phase saving
unless the atom is already assigned must-be-true, in which case true is chosen.
At every conflict for all atoms $b$ encountered in the (CDCL-style)
conflict analysis, the activity of $b$ is increased as follows: if $b$ represents a choice
point, its activity is increased; if $b$ is an ordinary atom the activity of all known choice points
that influence $b$ is increased.
This activity increment (initially 1) is divided by 0.92 after every conflict,
i.e., the increment increases with every conflict. This is a state-of-the-art
way of realizing the decay of activities by increasing the activity increment
instead. The relative order of activities stays the same as with the decaying,
but only the most recent value needs to be adapted instead of decaying all
activity values of all atoms.
Internally, atom activities are stored as double-precision floating point
values and whenever the activity of an atom exceeds $10^{100}$, all activities
are normalized (divided by $10^{100}$). The increment is also normalized.

\alphaslv's \emph{dependency-driven MOMs} implementation used to initialize atom activities is inspired by \clingo's implementation and also exploits \textit{dependencies} as described above.
When a new nogood is produced by the grounder, the activities of all choice points that have an influence on one of the literals in the new nogood are updated to their current MOMs value.

\subsection{Learned-clause Deletion}
\label{sec:techniques:deletion}

Conflict-driven learning, usually considered the most important technique for
SAT solving (and ASP solving), leads to many additional clauses being learned
during search. Since each learned clause must be stored, this increases the
clause database significantly during the runtime of a solver (in the order of
thousands of new clauses per second). However, the more clauses the clause
database contains, the more time is required for propagation, hence
propagation speed decreases with more clauses being present. This holds true
even in the presence of efficient propagation techniques like
two-watched-literals, or its adaption to the lazy-grounding setting
(cf.~\citeNP{DBLP:conf/inap/LeutgebW17}). Therefore, the learned-clause database
is regularly cleaned \cite{DBLP:conf/sat/EenS03}.

Some learned clauses must be excluded from being deleted, namely those that
are \emph{locked}, i.e., clauses that imply one of the currently assigned
literals. Since each learned clause helps to identify portions of the search
space where no solution (or answer set) can be found, deleting the wrong
clauses may increase the search space to consider as the solver
has to re-evaluate portions of the search space that otherwise would be
excluded by a learned clause. There are several ways to identify clauses that
are seemingly not important. The first is an \emph{activity} counter that is
incremented whenever a clause occurs in some conflict analysis, i.e., it
contributes to a conflict. Clauses with low activity are then deleted first as
they do not contribute much to the overall search performance. The second way
is to use the \emph{LBD} measure to determine a clauses quality
\cite{DBLP:conf/ijcai/AudemardS09}. Again, clauses with a poor LBD value (i.e.,
whose LBD is high) are deleted first. A combination of both is also common,
where activity is used to identify clauses for removal, but clauses with
exceptionally good LBD value are kept regardless.

\paragraph{Learned-clause deletion for lazy grounding.}
The technique of learned-clause (or nogood) deletion requires no special
adaptions to fit the lazy-grounding setting and we observed no particular
effects when realizing it in \alphaslv. The implementation in \alphaslv\ in general
mimics the default behaviour of \clingo\ and so clause database cleaning is
run after initial 2000 conflicts and that value increases by 100 for each
cleaning cycle. The whole sequence is reset after 20 cycles.

At each cleaning, half of the clauses are scheduled for removal. For that, the
average activity of the learned clauses is computed and $1.5$ times the
average is taken as threshold for removal, i.e., clauses with less than $1.5$
times the average activity are removed unless they are locked. Locked clauses are
not removed and as soon as half of the clause database has been removed the
process stops, keeping any remaining clauses even if their activity is below
the threshold. Note that this does not guarantee that half of the clauses are
actually removed, but it is a sufficiently good and efficiently computable
approximation. Note that clauses with a very good LBD value ($\leq 2$) are
never removed and they are not considered in the cleaning.

\section{Experimental Results}
\label{sec:experiments}

To asses the impact of newly adapted techniques in the lazy-grounding setting, we
evaluated them against six benchmark problems: Graph Colouring, House
Reconfiguration Problem (HRP), Stable Marriage,
Partner Units Polynomial (PUP), Non-Partition-Removal-Colouring (NPRC), and the
evaluation of nondeterministic L-Systems (Lindenmayer Systems).

\paragraph{Experimental Setup.}
Experiments were run on a cluster of machines each with two
Intel\textsuperscript{\textregistered} Xeon\textsuperscript{\textregistered} CPU E5-2650 v4 @ 2.20GHz with 12 cores each, 252 GB of memory, and Ubuntu 16.04.1 LTS Linux.
Benchmarks were scheduled with the ABC Benchmarking System \cite{DBLP:conf/aiia/Redl16} together with HTCondor\textsuperscript{\texttrademark}.\footnote{\url{https://github.com/credl/abcbenchmarking}, \url{http://research.cs.wisc.edu/htcondor}}
Time and memory consumption was measured by \slv{pyrunlim},\footnote{\url{https://alviano.com/software/pyrunlim/}}
which was also used to limit wall time consumption to 5 minutes per instance and swapping to 0.
\alphaslv\ was used in several configurations to compare the impact of different solving techniques.
In every configuration, constraints were grounded permissively and rules were grounded strictly as suggested by \citeN{DBLP:conf/lpnmr/TaupeWF19}.
When using \clingo\ 5.3.0, some techniques not yet supported by \alphaslv\ were switched off to improve comparability.\footnote{\clingo\ was used with the switches \texttt{--sat-prepro=no --eq=0}.}
For additional comparisons, \omiga\ \cite{omiga_system} was used in learning mode \cite{Weinzierl.2013}, and \asperix\ 0.2 \cite{Lefevre.2017} was used with command-line argument \texttt{-N 1000000}.
Moreover, \lazywasp\ \cite{DBLP:journals/tplp/CuteriDRS19}, which is a recent ground-and-solve system that incorporates powerful partial evaluation techniques, was used in its default configuration. Since \lazywasp\ requires a manual splitting of programs into a lazily evaluated part and a part that is evaluated with ground-and-solve techniques, the splitting was done such that a maximum part is evaluated lazily.
All systems were instructed to search for 10 answer sets.\footnote{Obtaining more than one (maybe trivial) answer set is often desirable. The number 10 has been chosen arbitrarily.}
The whole set of benchmarks was run three times; we report median solving times per instance in the discussion of the results below.

\paragraph{Encodings and Instances.}
The encodings for Graph Colouring and Stable Marriage were taken from the Fourth Answer Set Programming Competition \cite{Alviano.2013},
the former without modifications, the latter
with a choice rule replacing the equivalent disjunctive rule of the original.
The encoding for HRP was obtained from \citeN{DBLP:conf/confws/FriedrichRFHSS11} and adapted to conform to the input language of \alphaslv.\footnote{Currently, \alphaslv\ accepts only a subset of aggregate atoms and no optimization statements.}
The encoding for PUP was taken from the Third Answer Set Programming Competition \cite{aspcomp11}; choice rules have been used instead of disjunction.
Encoding and all 110 instances for NPRC were taken from \citeN{DBLP:conf/ijcai/BogaertsW18}.
For Graph Colouring and PUP, all instances from the ASP Competitions \cite{aspcomp11,Alviano.2013,Calimeri.2016} were used (60 for Graph Colouring, 65 for PUP).
For Stable Marriage, the 341 random instances generated by \citeN{DBLP:conf/lpnmr/TaupeWF19} were used again.
For HRP, the 47 instances generated for the ASP Challenge 2019 were used, which include instances of different problem classes and of varying difficulty and size.\footnote{\url{https://sites.google.com/view/aspcomp2019}}

The evaluation of nondeterministic L-Systems is a novel benchmark. L-Systems (or Lindenmayer Systems) are types of formal grammars, where production rules expanding symbols into larger sequences of symbols are applied to an initial starting word in parallel. Such generated words can be visualised using suitable drawing functions, resulting in fractal structures like, e.g.~the Cantor set, or a fractal tree. Each iteration of the evaluation of an L-System then typically yields one level of the resulting fractal structure. Our benchmark set is comprised of 39 instances of L-Systems. Some of them are deterministic, some nondeterministic (i.e., multiple different production rules may be applied to the same symbol). For the latter, additional constraints enforce global conditions on the generated words. Since words often grow exponentially with increased iteration steps, instances only compute few steps ($4$ to $20$) of the given L-System.

All encodings and instances as well as binaries of the \alphaslv\ version used for the experiments are available on our website.\footnote{\url{https://ainf.aau.at/dynacon/}}

\paragraph{Results and Discussion.}

\Cref{fig:cactus_GraphColouring,fig:cactus_House,fig:cactus_StableMarriage,fig:cactus_pup,fig:cactus_NPRC,fig:cactus_lsystems} show cactus plots for the time consumed to solve each of the six benchmark problems.
They have been created in the usual way, i.e., the x axis gives the number of
instances solved within real (i.e., wall-clock) time given on the y axis.
Solving time per instance is the median across three solver runs.
Note that the y axis shows time accumulated over all solved instances.
We compare the runtimes of \clingo, \lazywasp, \omiga, and \asperix\ to that of various \alphaslv\ configurations.
The baseline configuration of \alphaslv\ is its latest implementation before introduction of the solving techniques presented in this paper, with permissive lazy grounding of constraints and strict lazy grounding of rules as described by \citeN{DBLP:conf/lpnmr/TaupeWF19}.
This configuration has been included to be able to study the accumulated effect of various sets of newly introduced solving techniques, which constitute the other four configurations:
Each of those employs our dependency-driven form of VSIDS together with phase saving, where the default phase is true in two configurations and false in the other two, and restarts are switched on in two configurations and off in the others.

\begin{figure}
	\centering
	\begin{minipage}{.7\textwidth}
		\centering
		\resizebox{\linewidth}{!}{
\begingroup%
\makeatletter%
\begin{pgfpicture}%
\pgfpathrectangle{\pgfpointorigin}{\pgfqpoint{6.304083in}{2.587276in}}%
\pgfusepath{use as bounding box, clip}%
\begin{pgfscope}%
\pgfsetbuttcap%
\pgfsetmiterjoin%
\definecolor{currentfill}{rgb}{1.000000,1.000000,1.000000}%
\pgfsetfillcolor{currentfill}%
\pgfsetlinewidth{0.000000pt}%
\definecolor{currentstroke}{rgb}{1.000000,1.000000,1.000000}%
\pgfsetstrokecolor{currentstroke}%
\pgfsetdash{}{0pt}%
\pgfpathmoveto{\pgfqpoint{0.000000in}{0.000000in}}%
\pgfpathlineto{\pgfqpoint{6.304083in}{0.000000in}}%
\pgfpathlineto{\pgfqpoint{6.304083in}{2.587276in}}%
\pgfpathlineto{\pgfqpoint{0.000000in}{2.587276in}}%
\pgfpathclose%
\pgfusepath{fill}%
\end{pgfscope}%
\begin{pgfscope}%
\pgfsetbuttcap%
\pgfsetmiterjoin%
\definecolor{currentfill}{rgb}{1.000000,1.000000,1.000000}%
\pgfsetfillcolor{currentfill}%
\pgfsetfillopacity{0.800000}%
\pgfsetlinewidth{1.003750pt}%
\definecolor{currentstroke}{rgb}{0.800000,0.800000,0.800000}%
\pgfsetstrokecolor{currentstroke}%
\pgfsetstrokeopacity{0.800000}%
\pgfsetdash{}{0pt}%
\pgfpathmoveto{\pgfqpoint{0.038889in}{-0.000000in}}%
\pgfpathlineto{\pgfqpoint{6.265194in}{-0.000000in}}%
\pgfpathquadraticcurveto{\pgfqpoint{6.304083in}{-0.000000in}}{\pgfqpoint{6.304083in}{0.038889in}}%
\pgfpathlineto{\pgfqpoint{6.304083in}{2.548387in}}%
\pgfpathquadraticcurveto{\pgfqpoint{6.304083in}{2.587276in}}{\pgfqpoint{6.265194in}{2.587276in}}%
\pgfpathlineto{\pgfqpoint{0.038889in}{2.587276in}}%
\pgfpathquadraticcurveto{\pgfqpoint{0.000000in}{2.587276in}}{\pgfqpoint{0.000000in}{2.548387in}}%
\pgfpathlineto{\pgfqpoint{0.000000in}{0.038889in}}%
\pgfpathquadraticcurveto{\pgfqpoint{0.000000in}{-0.000000in}}{\pgfqpoint{0.038889in}{-0.000000in}}%
\pgfpathclose%
\pgfusepath{stroke,fill}%
\end{pgfscope}%
\begin{pgfscope}%
\pgfsetbuttcap%
\pgfsetroundjoin%
\pgfsetlinewidth{1.505625pt}%
\definecolor{currentstroke}{rgb}{0.121569,0.466667,0.705882}%
\pgfsetstrokecolor{currentstroke}%
\pgfsetdash{{9.600000pt}{2.400000pt}{1.500000pt}{2.400000pt}}{0.000000pt}%
\pgfpathmoveto{\pgfqpoint{0.077778in}{2.441443in}}%
\pgfpathlineto{\pgfqpoint{0.466667in}{2.441443in}}%
\pgfusepath{stroke}%
\end{pgfscope}%
\begin{pgfscope}%
\definecolor{textcolor}{rgb}{0.000000,0.000000,0.000000}%
\pgfsetstrokecolor{textcolor}%
\pgfsetfillcolor{textcolor}%
\pgftext[x=0.622222in,y=2.373387in,left,base]{\color{textcolor}\fontsize{14.000000}{16.800000}\selectfont naive}%
\end{pgfscope}%
\begin{pgfscope}%
\pgfsetrectcap%
\pgfsetroundjoin%
\pgfsetlinewidth{1.505625pt}%
\definecolor{currentstroke}{rgb}{0.549020,0.337255,0.294118}%
\pgfsetstrokecolor{currentstroke}%
\pgfsetdash{}{0pt}%
\pgfpathmoveto{\pgfqpoint{0.077778in}{2.160665in}}%
\pgfpathlineto{\pgfqpoint{0.466667in}{2.160665in}}%
\pgfusepath{stroke}%
\end{pgfscope}%
\begin{pgfscope}%
\definecolor{textcolor}{rgb}{0.000000,0.000000,0.000000}%
\pgfsetstrokecolor{textcolor}%
\pgfsetfillcolor{textcolor}%
\pgftext[x=0.622222in,y=2.092610in,left,base]{\color{textcolor}\fontsize{14.000000}{16.800000}\selectfont dependency-driven VSIDS with phase-saving (alltrue)}%
\end{pgfscope}%
\begin{pgfscope}%
\pgfsetrectcap%
\pgfsetroundjoin%
\pgfsetlinewidth{1.505625pt}%
\definecolor{currentstroke}{rgb}{1.000000,0.498039,0.054902}%
\pgfsetstrokecolor{currentstroke}%
\pgfsetdash{}{0pt}%
\pgfpathmoveto{\pgfqpoint{0.077778in}{1.868804in}}%
\pgfpathlineto{\pgfqpoint{0.466667in}{1.868804in}}%
\pgfusepath{stroke}%
\end{pgfscope}%
\begin{pgfscope}%
\definecolor{textcolor}{rgb}{0.000000,0.000000,0.000000}%
\pgfsetstrokecolor{textcolor}%
\pgfsetfillcolor{textcolor}%
\pgftext[x=0.622222in,y=1.800749in,left,base]{\color{textcolor}\fontsize{14.000000}{16.800000}\selectfont dependency-driven VSIDS with phase-saving (allfalse)}%
\end{pgfscope}%
\begin{pgfscope}%
\pgfsetbuttcap%
\pgfsetroundjoin%
\pgfsetlinewidth{1.505625pt}%
\definecolor{currentstroke}{rgb}{0.172549,0.627451,0.172549}%
\pgfsetstrokecolor{currentstroke}%
\pgfsetdash{{9.600000pt}{2.400000pt}{1.500000pt}{2.400000pt}}{0.000000pt}%
\pgfpathmoveto{\pgfqpoint{0.077778in}{1.576943in}}%
\pgfpathlineto{\pgfqpoint{0.466667in}{1.576943in}}%
\pgfusepath{stroke}%
\end{pgfscope}%
\begin{pgfscope}%
\definecolor{textcolor}{rgb}{0.000000,0.000000,0.000000}%
\pgfsetstrokecolor{textcolor}%
\pgfsetfillcolor{textcolor}%
\pgftext[x=0.622222in,y=1.508888in,left,base]{\color{textcolor}\fontsize{14.000000}{16.800000}\selectfont dependency-driven VSIDS with phase-saving (alltrue) plus restarts}%
\end{pgfscope}%
\begin{pgfscope}%
\pgfsetbuttcap%
\pgfsetroundjoin%
\pgfsetlinewidth{1.505625pt}%
\definecolor{currentstroke}{rgb}{0.839216,0.152941,0.156863}%
\pgfsetstrokecolor{currentstroke}%
\pgfsetdash{{1.500000pt}{2.475000pt}}{0.000000pt}%
\pgfpathmoveto{\pgfqpoint{0.077778in}{1.285082in}}%
\pgfpathlineto{\pgfqpoint{0.466667in}{1.285082in}}%
\pgfusepath{stroke}%
\end{pgfscope}%
\begin{pgfscope}%
\definecolor{textcolor}{rgb}{0.000000,0.000000,0.000000}%
\pgfsetstrokecolor{textcolor}%
\pgfsetfillcolor{textcolor}%
\pgftext[x=0.622222in,y=1.217027in,left,base]{\color{textcolor}\fontsize{14.000000}{16.800000}\selectfont dependency-driven VSIDS with phase-saving (allfalse) plus restarts}%
\end{pgfscope}%
\begin{pgfscope}%
\pgfsetbuttcap%
\pgfsetroundjoin%
\pgfsetlinewidth{1.505625pt}%
\definecolor{currentstroke}{rgb}{0.090196,0.745098,0.811765}%
\pgfsetstrokecolor{currentstroke}%
\pgfsetdash{{5.550000pt}{2.400000pt}}{0.000000pt}%
\pgfpathmoveto{\pgfqpoint{0.077778in}{1.002944in}}%
\pgfpathlineto{\pgfqpoint{0.466667in}{1.002944in}}%
\pgfusepath{stroke}%
\end{pgfscope}%
\begin{pgfscope}%
\definecolor{textcolor}{rgb}{0.000000,0.000000,0.000000}%
\pgfsetstrokecolor{textcolor}%
\pgfsetfillcolor{textcolor}%
\pgftext[x=0.622222in,y=0.934888in,left,base]{\color{textcolor}\fontsize{14.000000}{16.800000}\selectfont clingo}%
\end{pgfscope}%
\begin{pgfscope}%
\pgfsetbuttcap%
\pgfsetroundjoin%
\pgfsetlinewidth{1.505625pt}%
\definecolor{currentstroke}{rgb}{0.498039,0.498039,0.498039}%
\pgfsetstrokecolor{currentstroke}%
\pgfsetdash{{1.500000pt}{2.475000pt}}{0.000000pt}%
\pgfpathmoveto{\pgfqpoint{0.077778in}{0.729749in}}%
\pgfpathlineto{\pgfqpoint{0.466667in}{0.729749in}}%
\pgfusepath{stroke}%
\end{pgfscope}%
\begin{pgfscope}%
\definecolor{textcolor}{rgb}{0.000000,0.000000,0.000000}%
\pgfsetstrokecolor{textcolor}%
\pgfsetfillcolor{textcolor}%
\pgftext[x=0.622222in,y=0.661694in,left,base]{\color{textcolor}\fontsize{14.000000}{16.800000}\selectfont asperix}%
\end{pgfscope}%
\begin{pgfscope}%
\pgfsetbuttcap%
\pgfsetroundjoin%
\pgfsetlinewidth{1.505625pt}%
\definecolor{currentstroke}{rgb}{0.580392,0.403922,0.741176}%
\pgfsetstrokecolor{currentstroke}%
\pgfsetdash{{5.550000pt}{2.400000pt}}{0.000000pt}%
\pgfpathmoveto{\pgfqpoint{0.077778in}{0.458694in}}%
\pgfpathlineto{\pgfqpoint{0.466667in}{0.458694in}}%
\pgfusepath{stroke}%
\end{pgfscope}%
\begin{pgfscope}%
\definecolor{textcolor}{rgb}{0.000000,0.000000,0.000000}%
\pgfsetstrokecolor{textcolor}%
\pgfsetfillcolor{textcolor}%
\pgftext[x=0.622222in,y=0.390638in,left,base]{\color{textcolor}\fontsize{14.000000}{16.800000}\selectfont omiga}%
\end{pgfscope}%
\begin{pgfscope}%
\pgfsetrectcap%
\pgfsetroundjoin%
\pgfsetlinewidth{1.505625pt}%
\definecolor{currentstroke}{rgb}{0.890196,0.466667,0.760784}%
\pgfsetstrokecolor{currentstroke}%
\pgfsetdash{}{0pt}%
\pgfpathmoveto{\pgfqpoint{0.077778in}{0.185500in}}%
\pgfpathlineto{\pgfqpoint{0.466667in}{0.185500in}}%
\pgfusepath{stroke}%
\end{pgfscope}%
\begin{pgfscope}%
\definecolor{textcolor}{rgb}{0.000000,0.000000,0.000000}%
\pgfsetstrokecolor{textcolor}%
\pgfsetfillcolor{textcolor}%
\pgftext[x=0.622222in,y=0.117444in,left,base]{\color{textcolor}\fontsize{14.000000}{16.800000}\selectfont lazy wasp}%
\end{pgfscope}%
\end{pgfpicture}%
\makeatother%
\endgroup%
}
		Legend for \cref{fig:cactus_GraphColouring,fig:cactus_House,fig:cactus_StableMarriage,fig:cactus_pup,fig:cactus_NPRC,fig:cactus_lsystems}.
	\end{minipage}

	\vspace{\floatsep}

	\begin{minipage}{.45\textwidth}
		\centering
		\resizebox{\linewidth}{!}{\input{cactus_GraphColouring.pgf}}
		\caption{Time consumption on Graph Colouring.}
		\label{fig:cactus_GraphColouring}
	\end{minipage}%
	\hfill
	\begin{minipage}{.45\textwidth}
		\centering
		\resizebox{\columnwidth}{!}{\input{cactus_StableMarriage.pgf}}
		\caption{Time consumption on Stable Marriage.}
		\label{fig:cactus_StableMarriage}
	\end{minipage}
	
	\vspace{\floatsep}
	
	\begin{minipage}{.45\textwidth}
		\centering
		\resizebox{\columnwidth}{!}{\input{cactus_House.pgf}}
		\caption{Time consumption on HRP.}
		\label{fig:cactus_House}
	\end{minipage}%
	\hfill
	\begin{minipage}{.45\textwidth}
		\centering
		\resizebox{\columnwidth}{!}{\input{cactus_pup.pgf}}
		\caption{Time consumption on PUP.}
		\label{fig:cactus_pup}
	\end{minipage}
	
	\vspace{\floatsep}
	
	\begin{minipage}{.45\textwidth}
		\centering
		\resizebox{\columnwidth}{!}{\input{cactus_NPRC.pgf}}
		\caption{Time consumption on NPRC.}
		\label{fig:cactus_NPRC}
	\end{minipage}%
	\hfill
	\begin{minipage}{.45\textwidth}
		\centering
		\resizebox{\columnwidth}{!}{\input{cactus_lsystems.pgf}}
		\caption{Time consumption on L-Systems.}
		\label{fig:cactus_lsystems}
	\end{minipage}%
\end{figure}

\Cref{fig:cactus_GraphColouring} shows that for the first time,
\alphaslv\ is able to solve several hard instances from the
ASP competitions (here for the Graph Colouring problem). This is a
breakthrough since those instances are hand-picked to exercise search
techniques of ground-and-solve systems, even though \clingo\ and \lazywasp\ still outperform \alphaslv.
All configurations employing
additional solving techniques outperform \alphaslv's baseline. The best
configuration even outperforms the baseline by a factor of three, allowing it to
solve 12 instead of the previous 4 instances.  Restarts appear to be a
particularly useful improvement for this benchmark, which is in line with our
observation that restarts perform well if choices are not ``stacked'' on each
other, as is the case here.

As can be seen in \cref{fig:cactus_House}, \alphaslv\ also profits from the new solving techniques when solving HRP.
All novel configurations clearly outperform the baseline, solving more
instances than the baseline. None of the various
settings, however, clearly performs better than the others for these HRP instances.

On Stable Marriage (\cref{fig:cactus_StableMarriage}) no improvement can be
observed. In fact, the baseline performs best.
At the
moment we are not sure why this is the case, but \clingo's effortless
performance indicates that there may be some other techniques missing for the
lazy-grounding setting. This is also underscored by \lazywasp's performance which is similar to \clingo's, since both employ similar ground-and-solve techniques.

Many more PUP instances can be solved when employing the new solving techniques including restarts, even though they consume more time on easier instances compared to some configurations without restarts (\cref{fig:cactus_pup}). 

Note that all of the above problems are easy to ground, hence lazy grounding
is not necessary. We still picked those to demonstrate that the search
performance of lazy grounding is increasingly improving even on problems where
lazy grounding per se does not improve performance. Actually, the above
problems all present a worst-case scenario (i.e., compared to ground-and-solve
systems, a lazy-grounding system only lacks some information).

The fifth problem, NPRC, is  one where grounding itself is also an
issue. As shown in \cref{fig:cactus_NPRC}, \alphaslv\ clearly outperforms
\clingo\ on this problem. With regard to the novel techniques in \alphaslv, on the
one hand, there is some variance but no clear improvement over the baseline however, on the other
hand, this indicates that the novel techniques help to solve hard search
problems while introducing no obstacles for solving hard-to-ground
instances. The \lazywasp\ system also performs significantly better than \clingo\ and comparably to \alphaslv. We also noted that \lazywasp's runtime varies wildly even when run repeatedly on the same instance. We currently have no explanation for this behaviour and guess it might be due to some randomization.

\Cref{fig:cactus_lsystems} shows the results for evaluating nondeterministic L-Systems. This benchmark is grounding-intense, so \clingo\ can only solve the easier instances and partial evaluation techniques of \lazywasp\ have no positive effect.
\alphaslv\ is able to solve most instances and there is a clear distinction between those configurations with the initial phase being true and those with false, as the latter are only able to solve the most simple instances. Whether restarting is enabled or not seems to make little difference. Both the baseline and the configuration with restarts and dependency-driven VSIDS solve the same number of instances, though the latter needs a bit more time.
No line for \asperix\ is visible, but it is able to solve the smallest instance.\footnote{This benchmark could not be run with \omiga\ as a bug prevents it from solving such ASP encodings.}

In all figures, only few data points can be seen for \omiga\ and \asperix, because those systems could only solve very few instances.
Furthermore, HRP was not used with \omiga\ and \asperix\ because of the restricted input languages of these systems, and \omiga\ produced several exceptions when trying to solve Stable Marriage instances.

Overall, adapting restarts, phase saving, dependency-driven VSIDS and
learned-clause deletion to the lazy-grounding setting is a significant
improvement for lazy-grounding ASP solving. It improves search performance on
hard problems, sometimes dramatically, and still allows the
grounding bottleneck to be avoided.

\section{Related Work}
\label{sec:related}

There are several approaches to tackle the grounding bottleneck of
ASP. The grounders of ground-and-solve systems have, for a long time, been trying to
minimize the size of the resulting ground program, which gave rise to
intelligent grounding techniques
(cf.~\citeNP{DBLP:conf/lpnmr/GebserKKS11}; \citeNP{Leone.2006}; \citeNP{DBLP:journals/ia/CalimeriFPZ17}).

A more recent attempt to circumvent the grounding bottleneck is by extending
ASP with specific problem solvers (e.g.\ temporal
\cite{DBLP:conf/lpnmr/CabalarKMS19}, or difference-logic
\cite{DBLP:conf/lpnmr/AbelsJOSTW19} reasoners)
\cite{DBLP:conf/iclp/GebserKKOSW16} and then manually reformulating part of
the original problem in the added formalism. Besides the need to develop and
integrate the specific problem solvers, it requires users of ASP to be
knowledgeable in another (unrelated) formalism to solve their problems.

Another approach aims to tackle the grounding issue by grounding only those
parts of a first-order theory which are actually needed to solve the problem
at hand. Several techniques follow this general idea. Incremental grounding
\cite{DBLP:journals/aicom/GebserSS11}, which works for planning and related
types of problems, introduces time steps on-the-fly when the solver notices
no solution exists in the given time window.
Partial compilation techniques \cite{DBLP:journals/tplp/CuteriDRS19} are a
recent approach, where a stratifiable part of the program is automatically
turned into a lazy propagator. This successfully addresses the grounding
bottleneck for ASP programs with a certain structure, as also shown by our
experiments. It currently requires the user to manually identify the program
part that can be turned into a lazy propagator, however.  Also top-down
lazy-model generation \cite{DBLP:journals/jair/CatDBS15} and top-down stable
model generation techniques
\cite{DBLP:conf/ppdp/MarpleBMG12,DBLP:journals/corr/abs-1709-00501} exist. The
former, however, does not work on ASP but the related formalism of FO(ID),
while the latter, to the best of our knowledge, does not achieve good solving
efficiency.

The first bottom-up lazy-grounding systems available were GASP \cite{gasp} and
ASPeRiX \cite{Lefevre.2017}. The \omiga\ solver \cite{omiga_system} uses a
Rete-network for efficient grounding and propagation, but, like its
predecessors, does not provide efficient search.

Part of the previous work in \alphaslv\ focused on the formulation and
integration of domain-specific heuristics to solve large-scale instances where
such heuristics are known
(cf.~\citeNP{DBLP:journals/corr/abs-1909-08231}). The introduction of
domain-independent state-of-the-art techniques employed for grounded ASP
solving, however, was left open until now.

\section{Conclusions and Future Work}
\label{sec:conclusion}
Lazy-grounding ASP solvers must address the grounding bottleneck whilst providing
problem solving techniques which allow the
solution of hard problem instances. Problem solving techniques which proved to
be successful for grounded ASP programs cannot be directly transferred to
lazy-grounding solvers. In this paper we reviewed various problem solving
techniques such as restarts, phase saving, domain-independent
heuristics,\footnote{Note that we focus here on heuristics for the
  \textit{solver} component only, while heuristics for the \textit{grounder}
  in a lazy-grounding system are subject of future work.} and learned-clause
deletion. We presented enhancements and adaptations such that these techniques
are applicable in lazy-grounding ASP solvers.

Experimental analysis on the \alphaslv\ solver showed significant improvements
(up to a factor of three) on some hard instances while for other problems the
additional techniques have no negative effect. Similarly, as for other solvers,
\alphaslv\ comes now with a range of search options and there does not seem to
be a setting that is always preferable. Hence portfolio solving might improve
efficiency further.

As regards future work, we want to investigate further improvements to
current solving techniques, like blocking restarts in certain
cases. Furthermore, integrating external atoms similar to those by
\citeN{DBLP:journals/jair/EiterKRW18} is another goal.

\subsection*{Acknowledgments}
This work has been conducted in the scope of the research project \textit{DynaCon (FFG-PNr.:\ 861263)}, which is funded by the Austrian Federal Ministry of Transport, Innovation and Technology (BMVIT) under the program \enquote{ICT of the Future} between 2017 and 2020,\footnote{See \url{https://iktderzukunft.at/en/} for more information.} and in the scope of the research project Productive4.0, which is funded by EU-ECSEL under grant agreement no737459.
The first author would like to thank Bart Bogaerts and Jakob
Nordstr{\"o}m for the discussions of modern SAT techniques.

\pagebreak

\bibliographystyle{acmtrans}
\bibliography{bibliography}
\end{document}